# Multiobjective Optimization of Solar Powered Irrigation System with Fuzzy Type-2 Noise Modelling

**T. Ganesan***
*Department of Chemical Engineering,
Universiti Teknologi Petronas,
32610 Bandar Seri Iskandar, Malaysia.
*email: tim.ganesan@gmail.com*

**P. Vasant**
*Department of Fundamental & Applied Sciences,
Universiti Teknologi Petronas,
32610 Bandar Seri Iskandar, Malaysia.*

**I. Elamvazuthi**
*Department of Electrical & Electronics Engineering,
Universiti Teknologi Petronas,
32610 Bandar Seri Iskandar, Malaysia.*

**ABSTRACT**

*Optimization is becoming a crucial element in industrial applications involving sustainable alternative energy systems. During the design of such systems, the engineer/decision maker would often encounter noise factors (e.g. solar insolation and ambient temperature fluctuations) when their system interacts with the environment. In this chapter, the sizing and design optimization of the solar powered irrigation system was considered. This problem is multivariate, noisy, nonlinear and multiobjective. This design problem was tackled by first using the Fuzzy Type II approach to model the noise factors. Consequently, the Bacterial Foraging Algorithm (BFA) (in the context of a weighted sum framework) was employed to solve this multiobjective fuzzy design problem. This method was then used to construct the approximate Pareto frontier as well as to identify the best solution option in a fuzzy setting. Comprehensive analyses and discussions were performed on the generated numerical results with respect to the implemented solution methods.*

## INTRODUCTION

Diesel generators, gas turbines and other fossil-fuel based power systems have been widely employed for driving conventional irrigation systems. Currently, various issues related to the utilization of fossil fuel in power systems have surfaced (e.g. price fluctuations, environmental concerns and efficiency). Hence in recent studies, the harnessing of solar energy to power irrigation pumps have become popular (Helikson *et al*., 1991; Wong & Sumathy, 2001; Jasim *et al.*, 2014; Dhimmar *et al*., 2014). The design and sizing of solar power systems greatly impacts the system's reliability, emissions and efficiency (Al-Ali *et al*., 2001). Thus challenges in effective sizing of this system falls into the realm of optimization. Here the optimal sizing and design are obtained such that the following targets are achieved (Shivrath *et al*., 2012; Carroqino *et al*.,2009):

- Increased savings
- Low emissions
- Good system efficiency
- High power output

Efforts in the optimization of solar powered irrigation systems have been performed via the implementation of metaheuristics such as: genetic algorithms (GA) and particle swarm optimization (PSO) (Gouws & Lukhwareni, 2012). However, due to the complexity of the system, the design needs to be carried out in such a way that it takes into account multiple aims simultaneously (i.e. multiobjective optimization) (Chen *et al*., 1995).

Real-world optimization problems often contain large degrees of uncertainties. To effectively handle such problems, higher order fuzzy logic (FL) such type-2 FL approaches are often employed in tandem with optimization techniques (Castillo and Melin, 2012; Ontoseno *et al*., 2013; Sánchez *et al*., 2015). Most existing research works involving the application of type-2 FL systems revolve around control theory and control system design (Fayek *et al*., 2014; Martinez *et al*., 2011; Bahraminejad *et al*., 2014; Oh *et al*., 2011; Linda and Manic, 2011). For instance in Wu and Tan (2004), the authors investigated the effectiveness of evolutionary type-2 FL controllers for uncertainty modelling in liquid-level processes. In that work, the authors employed the genetic algorithm (GA) to evolve the type-2 FL controller. This approach was found to perform very well for modelling uncertainties in complex plants compared to conventional type-1 FL frameworks. In Bahraminejad *et al*., (2014), a type-2 FL controller was employed for pitch control in wind turbines. Pitch control in wind turbines are critical for power regulation and reduction of fatigue load in the components of the turbine. In Bahraminejad *et al*., (2014), the type-2 FL controller was shown to significantly improve the adjustment of the pitch angle, rotor speed and power output of the wind turbine generator. Similarly, in Allawi (2014), a type-2 FL controller was utilized for controlling robots involved in cooperation and target achieving tasks in multi-robot navigation systems. In that work, the controllers were optimized using the Particle Swarm Optimization (PSO) and the Hybrid Reciprocal Velocity Obstacles techniques. The author discovered that the optimized type-2 FL controller performed very well for controlling such robots.

Besides control theory and engineering, type-2 FL has also been employed for modelling systems endowed with high levels of uncertainty. For instance in Paricheh and Zare (2013), a type-2 FL system was used to predict long-term traffic flow volume. In Paricheh and Zare (2013), the traffic flow data was heavily influenced by various time-dependent uncertainties and nonlinearities. In that work, the authors employed a type-2 FL system in combination with genetic algorithm and neural net-based approaches. Another interesting implementation of type-2 FL was presented in the work of Qiu *et al*., (2013). In that research, the authors focused on developing a general interval type-2 fuzzy C-means algorithm. The proposed fuzzy algorithm was employed for medical image segmentation in magnetic-resonance images (MRI). These MRIs are usually noisy and highly inhomogeneous. The works of Castillo and Melin (2012) and Dereli *et al*., (2011) provides a more comprehensive review on type-2 FL systems applied in industrial settings.

Swarm intelligence (SI) stands as one of the most favored strategies for solving complex optimization problems due to its effectiveness during search operations and in terms of computational cost (Liu *et al*., 2002). Some of the most popular SI-based techniques are, cuckoo search (CS) (Yildiz, 2013), ant colony optimization (ACO) (El-Wahed, 2008), PSO (Kennedy and Eberhart, 1995) and bacterial foraging algorithm (BFA) (Passino, 2002). In many past research works, PSO has been implemented extensively for solving nonlinear optimization problems. Recently, alternative optimization strategies such as BFA has become attractive to be implemented for such purposes. BFA's computational performance has been demonstrated to be as good as and sometimes better than other SI-based techniques (Al-Hadi and Hashim, 2011). BFA is inspired by the natural behavior of the E. Coli bacterium. This behaviour involves the search mechanisms utilized by the bacteria during nutrient foraging. These mechanism were then employed by Passino (2002) to design the BFA for solving complex optimization problems. The central principle of the BFA framework is as follows:

*'Each bacteria in the swarm tries to maximize its energy per unit time spent during the foraging for nutrients while simultaneously evading noxious substances'*

In recent times, BFA has been seen applied in many engineering applications (e.g. economic dispatch, engineering design, manufacturing technology, power systems and control systems). In the research work by Mezura-Montes and Hernandez-Ocana (2009), specific adjustments to the BFA approach was performed to enhance its optimization capability. This enhanced-BFA was then effectively implemented for engineering design. In Mezura-Montes *et al*., (2014), the design optimization of a crank-rocker-slider (variable transmission) system was performed using BFA.

The design optimization problem in Mezura-Montes *et al*., (2014) was formulated in two forms (single-objective and bi-objective). The authors then restructured the conventional BFA for multiobjective optimization. The conventional and multiobjective techniques were successfully implemented to optimize the mechanical design of the crank-rocker-slider system. BFA has also been applied in cellular manufacturing systems (Nouri and Hong, 2013). In these systems, the cell formation issue is tackled while considering the number of exceptional elements and cell load variations. In Nouri and Hong (2013), using the BFA approach, part families and machine cells were generated by the authors. In Panda *et al*., (2009), the BFA was seen to aid the manufacturing process of rapid prototyping. In that work, the BFA was used for optimizing the process parameters employed for fused deposition modeling (FDM).

Besides manufacturing, BFA has also been widely applied by engineers/researchers in power engineering and distribution. In such applications the BFA is utilized for optimizing economic load dispatch. In economic load dispatch, the target is to obtain the most optimal load dispatch for the power generating units while taking into account variable load demands and load constraints (Vijay, 2012). In addition, BFA has been utilized for obtaining optimal power flow (i.e. economic and efficient) in flexible alternating current transmission system (FACTS) devices (Ravi *et al*., 2014).

This chapter is organized as follows: Section 2 presents an overview of type-2 fuzzy logic while Section 3 discusses some fundamentals involving the application of type-2 fuzzy logic for noise modeling. Section 4 contains information on solar-powered irrigation system design followed by the BFA approach give in Section 5. Section 6 provides some details regarding the sigma diversity metric employed in this work while Section 7 presents the computational results and analyses. Finally, this chapter ends with some conclusions and recommendations for future research works.

## OVERVIEW OF TYPE-2 FUZZY LOGIC

Type-2 fuzzy sets are generalizations of the conventional or type-1 fuzzy sets (Zadeh, 1975). The primary feature of the Type-1 fuzzy set is its membership function, $\eta_F(x) \in [0,1]$ and $x \in X$. Type-2 FL employs a membership function of a second order, $\mu_F(y, \eta_F(x)) \in [0,1]$ such that $y \in Y$. Therefore, $\mu_F(y, \eta_F(x))$ is a membership function that requires three-dimensional inputs. The type-2 fuzzy set is defined as follows:

$$\widetilde{F} = \{(y, \eta_F(x)), \mu_F(y, \eta_F(x)) : \forall x \in X, \forall y \in Y, \eta_F(x) \in [0,1]\} \qquad (1)$$

The type-2 membership function has two membership grades: primary and secondary memberships. Thus, a crisp set (or function) undergoes fuzzification twice such that the first fuzzification transforms it to a type-1 fuzzy set (via the primary membership function). Using the secondary membership function the type-1 fuzzy set is transformed to a type-2 fuzzy set. In simple terms, type-2 fuzzy set results from the fuzzification of a type-1 fuzzy set. This operation aims to improve its efficacy and accuracy in capturing uncertainties. The region covered by the type-1 fuzzy sets in type-2 FL systems is represented by the footprint of uncertainty (FOU). This region of uncertainty is contained by the uppermost and lowermost type-1 membership functions $\eta_F^U(x)$ and $\eta_F^L(x)$ respectively. A type-2 FL system usually consists of four subcomponents: fuzzifier, inference engine, type reducer and defuzzifier. The fuzzifier directly transforms the crisp set into a type-2 fuzzy set. The inference engine functions to combine rules to map the type-2 fuzzy set from crisp inputs. Therefore each rule is interpreted as a type-2 fuzzy implication in the inference engine. In this work, all the consequent and antecedent sets are generalized type-2 fuzzy sets. The rule, $R_i$ from a type-2 FL system could be generally represented as follows:

$$R_i : \text{IF } x_1 \text{ is } \widetilde{M}_1 \text{ AND}\ldots\text{AND } x_j \text{ is } \widetilde{M}_j \text{ THEN } y_1 \text{ is } \widetilde{N}_1, \ldots, y_k \text{ is } \widetilde{N}_k \text{ such that } i \in [1, Z] \qquad (2)$$

where $j$ is the number of fuzzy inputs, $k$ is the number of fuzzy outputs and $i$ is the number of rules. The type reducer functions to transform (or reduce) the type-2 fuzzy set to a type-1 fuzzy set. Various type-reduction approaches have been developed in the past. For instance: centroid type reduction (Mendel and John, 2002) vertical slice-centroid type reduction (Lucas *et al.*, 2007), alpha cuts/planes (Hamrawi and Coupland, 2009) and the random sampling technique (Greenfield *et al.*, 2005). Defuzzification on the other hand reduces the type-1 fuzzy set to a crisp output similar to operations in conventional type-1 FL systems. There are various defuzzification techniques which are employed selectively to suit specific data representations and applications (Rao and Saraf, 1996).

## TYPE-2 FUZZY LOGIC FOR NOISE MODELLING

Energy systems that rely on their surroundings are often difficult to design if the surroundings are noisy. In solar-powered systems, weather-dependent variables such as insolation and ambient temperature are often found to be irregular and noisy. Thus, when designing such systems, the model utilized should be able to account for such irregularities. In this work, equipped with meteorological data, type-2 FL is employed to model and incorporate insolation and ambient temperature into the optimization formulation. The meteorological data for Santa Rosa Station at California was retrieved from the weather database of the University of California Agriculture and Natural Resources. The daily average insolation (in W/m$^2$) and ambient temperature (in K) was obtained for every month of the year 2014. Table 1 provides the monthly ambient temperature and insolation data:

Table 1: Monthly average ambient temperature and insolation taken at Santa Rosa Station, California in 2014

| Month, $m$ | Ambient Temperature, $Z_a$ (K) | | | Average Solar Radiation, $Z_b$ (W/m$^2$) | | |
|---|---|---|---|---|---|---|
| | MAX | MIN | AVERAGE | MAX | MIN | AVERAGE |
| 1 (Jan) | 297.4 | 265.2 | 281.3 | 146 | 43 | 104.71 |
| 2 (Feb) | 295.8 | 266.3 | 281.05 | 186 | 16 | 107.11 |
| 3 (Mac) | 301.9 | 272.4 | 287.15 | 236 | 41 | 169.16 |
| 4 (April) | 304.7 | 273.6 | 289.15 | 295 | 77 | 240.03 |
| 5 (May) | 305.8 | 275.2 | 290.5 | 335 | 30 | 286.52 |
| 6 (Jun) | 306.3 | 276.3 | 291.3 | 336 | 211 | 306.20 |
| 7 (July) | 308 | 279.1 | 293.55 | 330 | 102 | 255.16 |
| 8 (Aug) | 306.9 | 279.7 | 293.3 | 282 | 62 | 220.71 |
| 9 (Sept) | 308 | 278.6 | 293.3 | 252 | 92 | 196.13 |
| 10 (Oct) | 309.1 | 273.6 | 291.35 | 216 | 27 | 149.87 |
| 11 (Nov) | 301.3 | 271.9 | 286.6 | 146 | 18 | 99.67 |
| 12 (Dec) | 293.6 | 270.8 | 282.2 | 115 | 14 | 62.45 |

The primary membership function, $\eta_F(x)$ was employed to model the monthly data while the secondary membership function, $\mu_F(y, \eta_F(x))$ was used to model the overall annual data. This way the noisy monthly fluctuations in the data is taken into consideration via the type-2 FL approach. The overall type-2 fuzzy modelling strategy is given in Figure 1:

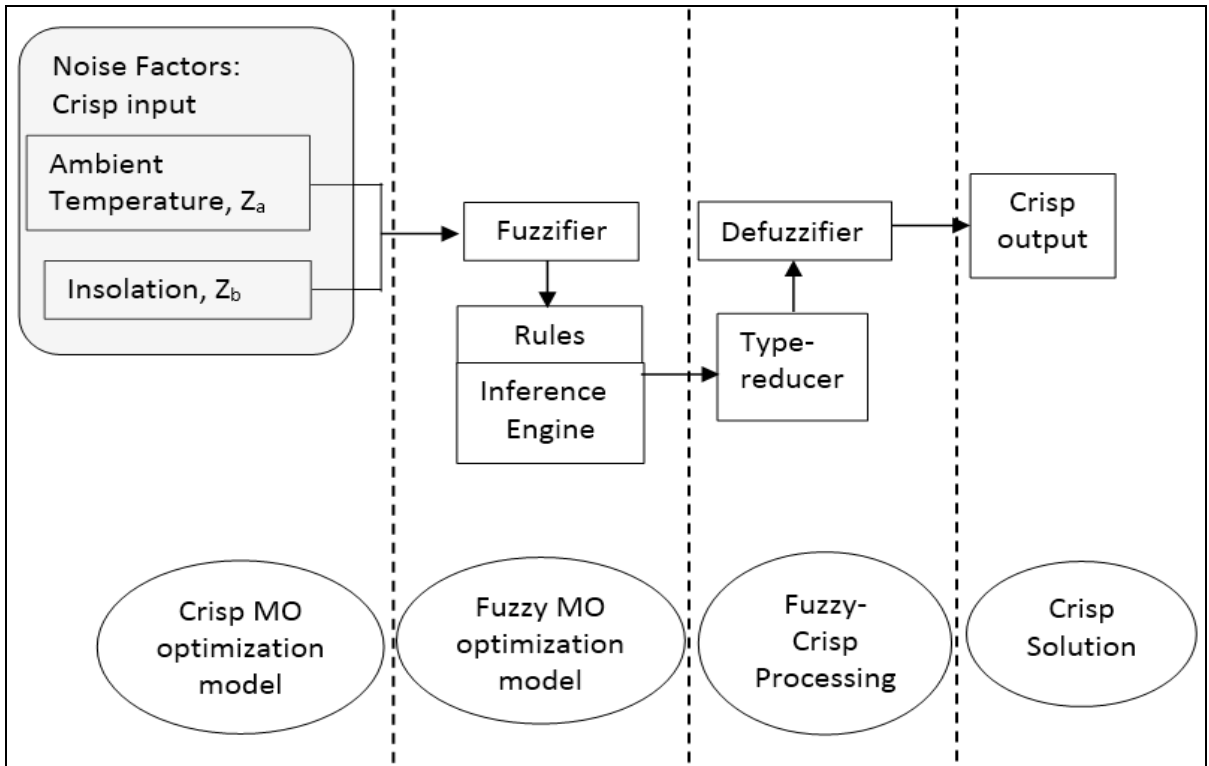

Figure 1: Type2 fuzzy modelling strategy

The S-curve function is employed as the primary and secondary membership functions ($\eta_F(x)$ and $\mu_F(y, \eta_F(x))$). Therefore type-2 fuzzification is performed on the ambient temperature ($Z_a$) and insolation, ($Z_b$) using the S-curve membership function. This is carried out by determining the average, maximum and minimum values of insolation and ambient temperature from the meteorological data. The S-curve membership function is as follows:

$$\mu_{\tilde{b}_i} = \begin{cases} 1 & \text{if } b_i \leq b_i^a \\ \dfrac{B}{1 + Ce^{\alpha\left(\frac{b_i - b_i^a}{b_i^b - b_i^a}\right)}} & \text{if } b_i^a \leq b_i \leq b_i^b \\ 0 & \text{if } b_i \geq b_i^b \end{cases} \quad (3)$$

where $B$ and $C$ are parameters which are tuned heuristically such that the membership fits the meteorological data effectively. Using Zadeh's extension principle, all crisp variables (ambient temperature ($Z_a$) and insolation ($Z_b$)) and their respective constraints are transformed via type-2 fuzzification. Assuming a credibility level $\varepsilon$, ($0 < \varepsilon < \frac{B}{1+C}$) chosen by the Decision Maker (DM), as he/she takes a risk and ignores all the membership degrees smaller than the $\varepsilon$ levels (Rommelfanger, 1989). The FOU is the union of all the primary memberships (Mo *et al.*, 2014). In this case, the union of all the primary S-curve memberships, $\eta_F(x)$ for each month depicts the FOU:

Let $\eta_F^i(x) \in \left(J_x^i \subseteq [0,1]\right)$ such that $i = [L, U]$, THEN
$$FOU = \bigcup_{x \in X} J_x^i \quad (4)$$

where $J_x^i$ is the fuzzy set, $L$ is the lower bound and $U$ is the upper bound. A graphical depiction of the FOU generated by the primary S-curve memberships in this work is given in Figure 2:

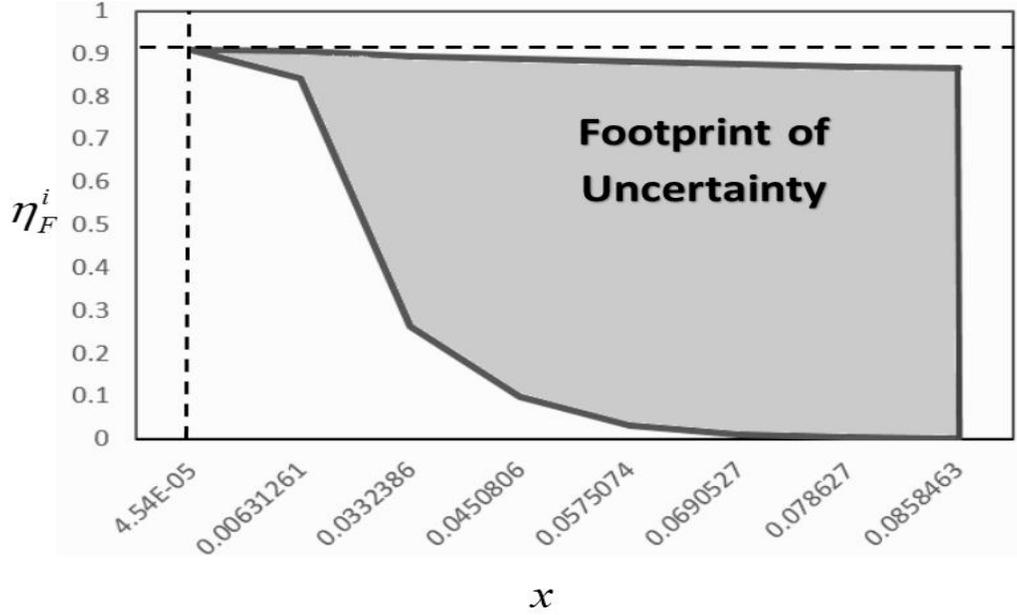

Figure 2: FOU generated by the primary S-curve memberships

There are many readily available techniques for type-reduction and defuzzification. In this work, the alpha-plane approach (Hamrawi and Coupland, 2009; Liu 2006) was employed for type-reduction while the conventional alpha-cut approach (Klir and Yuan, 1995) was used for the defuzzification. An alpha cut can be defined on a fuzzy set, $\tilde{F}$ via its decomposed form as follows:

$$\tilde{F} = \bigcup_{\alpha \in [0,1]} \alpha \cdot F_{\alpha} \qquad (5)$$

where $F_{\alpha}$ is an $\alpha$-level set. Similarly, an alpha-cut on a type-2 fuzzy set could performed via the decomposition theorem. Since this operation is performed on a type-2 fuzzy set, it is defined as an alpha-plane instead of an alpha cut:

$$\tilde{\tilde{F}} = \bigcup_{\tilde{\alpha} \in [0,1]} \tilde{\alpha} \cdot \tilde{F}_{\tilde{\alpha}} \qquad (6)$$

where $\tilde{F}_{\tilde{\alpha}}$ is a type-2 $\alpha$-level set. It should be noted that by using Zadeh's extension principle, the alpha-planes could be utilized to execute type-2 fuzzy operations using interval type-2 fuzzy sets. This is analogous to implementations in type-1 fuzzy sets since the extension principle could be evoked to extend functions that interrelate crisp, type-1 fuzzy as well as type-2 fuzzy sets.

## SOLAR-POWERED IRRIGATION SYSTEM

The design of the solar-powered irrigation system significantly affects its system characteristics. In Chen *et al.* (1995), the system characteristics which were given emphasis were the pump load/power output, $f_1$ (kW), overall efficiency, $f_2$ (%) and the fiscal savings, $f_3$ (USD). The design variables were: the maximum pressure, $x_a$ (MPa), maximum temperature, $x_b$ (K), maximum solar collector temperature, $x_c$ (K), the fluid flowrate, $x_d$ (kg/s).

Besides design parameters, noisy environmental factors such as the ambient temperature, $Z_a$ (K) and the level of insolation, $Z_b$ (W/m$^2$) greatly influences the system's characteristics. The objective functions and constraints are thus formulated such that they consider the design variable while taking into account the environmental factors. The design formulation is given as follows:

$$f_1 = -(24.947 + 16.011x_d + 1.306x_b + 0.820x_b x_d - 0.785Z_a - 0.497x_d Z_a + 0.228x_a x_b \\ + 0.212x_a - 0.15x_b^2 + 0.13x_a x_d - 0.11x_a^2 - 0.034x_b Z_a + 0.002x_a Z_a)10^{3.24} \quad (7)$$

$$f_2 = -43.4783(018507 + 0.01041x_a + 0.0038Z_b - 0.00366Z_a - 0.0035x_c - 0.00157x_b) \quad (8)$$

$$f_3 = -(174695.73 + 112114.69x_d + 9133.8x_b + 5733.05x_b x_d - 5487.76Z_a - 3478.84x_d Z_a \\ + 1586.48x_a x_b + 1486.84x_a - 1067.42x_b^2 + 916.26x_a x_d - 768.9x_a^2 - 242.88x_b Z_a \\ + 152.4x_a Z_a)10^{3.23} \quad (9)$$

$$x_a \in [0.3, 3]$$
$$x_b \in [450, 520]$$
$$x_c \in [520, 800]$$
$$x_d \in [0.01, 0.2]$$
$$Z_a \in [293, 303]$$
$$Z_b \in [800, 1000] \quad (10)$$

As mentioned in previous sections, to accurately account for the uncertainties arising from the noisy environment, the fuzzy type-2 approach is incorporated into the system design. Taking this view, the ambient temperature, $Z_a$ (K) and the level of insolation, $Z_b$ (W/m$^2$) is fuzzified. By implication, the target objectives and the constraints that bound the environmental factors ($Z_a$ and $Z_b$) in the model above is transformed to a fuzzy form. The type-1 fuzzy constraints for the environmental factors are as follows:

$$\tilde{Z}_a^m \in [Z_{a,\min}^m, Z_{a,\max}^m]$$
$$\tilde{Z}_b^m \in [Z_{b,\min}^m, Z_{b,\max}^m] \quad (11)$$

where $m$ represents the months in the year as in Table 1 while $Z_{a,\max}^m$ and $Z_{b,\max}^m$ are the maximum monthly average and $Z_{b,\max}^m$ and $Z_{b,\max}^m$ are the minimum monthly average. The type-2 fuzzy constraint for the whole year is represented as follows:

$$\tilde{\tilde{Z}}_a \in [\tilde{Z}_{a,\min}, \tilde{Z}_{a,\max}] = [265.2, 309.1]$$
$$\tilde{\tilde{Z}}_b \in [\tilde{Z}_{b,\min}, \tilde{Z}_{b,\max}] = [14, 336] \quad (12)$$

Therefore, the MO design optimization of the solar-powered irrigation system is effectively transformed to a Fuzzy MO optimization problem:

$$Maximize \rightarrow (\tilde{\tilde{f}}_1, \tilde{\tilde{f}}_2, \tilde{\tilde{f}}_3)$$

subject to

*Crisp Design Constraints & Type-2 Fuzzy Environmental Constraints* (13)

## BACTERIA FORAGING ALGORITHM

The dynamics of bacteria foraging is directly influenced by evolutionary biology. Thus, bacteria with successful foraging strategies would stand a better chance in propagating their genetic makeup as compared to bacteria with poor strategies. This way bacteria at successive generations always contain improved foraging strategies relative to past generations and the strategies continually improves as they go along reproducing. Due to such progressive behavior, many researches were targeted to model bacteria foraging dynamics as an optimization process. The central theme of foraging viewed from this perspective is that the organisms conduct the search in such a way that they maximize the energy they obtained from the nutrients at minimal time during foraging.

Foraging efforts vary according to the species of the organism and the environment where the foraging is taking place. For instance, herbivores would find it easier to locate food as compared to carnivores in any habitat. As for the environmental factor, the distribution of nutrients in desert or tundra conditions are sparser in contrast with the nutrient-rich tropical jungles. Design of effective and efficient search strategies for nutrient foraging which respects the previous constraints is critical for the long-term well-being of any organism. Another important factor to be considered for the design of effective search strategies is the type of nutrient. The type of nutrient will influence the fractionalization and planning of the strategy (O'Brien *et al*., 1990). For instance, consider a case where the nutrient is stationary but hidden in a hard shell (e.g. eggs). Then the organism would have design the foraging strategy in such a way that it searches for the shell (1), evades the nutrients parent (2), breaks the shell (3), consumes the nutrient (4) then escapes the nutrient location or nest before it gets attacked/killed (5).

In many organisms, synergetic foraging strategies are observed to emerge in nature (e.g. ants, bees and thermites). These organisms create communication mechanisms that enable them to share information about the foraging efforts led by each individual in the group. Such mechanisms provide the capability of the organisms to conduct 'group/swarm foraging'. Group foraging provides these organisms with a plethora of advantages such as increased protection against predators and enhanced hunting/foraging strategies. These advantageous traits increases the organism's chances for finding nutrients in good time. Besides synergetic strategies for foraging, other strategies such as cooperative building (Turner, 2011), group defense (Schneider and McNally, 1992) and other cooperative group behaviors are common in nature.

In the BFA, four main levels of loops are present in the technique (chemotaxis, swarming, reproduction and elimination-dispersal loops). These loops manage the main functional capabilities of the BFA. Each of the mentioned loops are designed according to bacteria foraging strategies and principles from evolutionary biology. These loops are executed iteratively until the total number of iterations, $N_T$ is satisfied. Each of the main loops may be iterated until some fitness condition is satisfied or until a user-defined loop cycle limit (chemotaxis ($N_c$), swarming ($N_s$), reproduction ($N_r$) and elimination-dispersal ($N_{ed}$)) is reached. In chemotaxis, the bacteria with the use of its flagellum, swims and tumbles towards the nutrient source. The tumbling mode allows bacterium motion in a fixed direction while the tumbling mode enables the bacterium to augment its search direction accordingly. Applied in tandem, these two modes gives the bacterium capability to stochastically move towards a sufficient source of nutrient. Thus, computationally chemotaxis is presented as follows:

$$\theta^i(j+1,k,l,m) = \theta^i(j,k,l,m) + C(i)\frac{\Delta(i)}{\sqrt{\Delta(i)\Delta^T(i)}} \quad (14)$$

where $\theta^j(j+1,k,l,m)$ is the $i^{th}$ bacterium at the $j^{th}$ chemotactic step, $k^{th}$ swarming step and $l^{th}$ reproductive step and $m^{th}$ elimination-dispersal step. $C(i)$ is the size of the step taken in a random direction which is fixed by the tumble, and $\Delta \in [-1,1]$ is the random vector.

In the swarming phase, the bacterium communicates to the entire swarm regarding the nutrient profile it mapped during its movement. The communication method adopted by the bacterium is cell-to-cell signaling. In *E.Coli* bacteria, aspartate is released by the cells if it is exposed to high amounts of succinate. This causes the bacteria to conglomerate into groups and hence move in a swarm of high bacterial density. The swarming phase is mathematically presented as follows:

$$J(\theta,P(j,k,l,m)) = \sum_{i=1}^{S}\left[-D_{att}\exp(-W_{att}\sum_{m=1}^{P}(\theta_m - \theta_m^i)^2)\right] + \sum_{i=1}^{S}\left[-H_{rep}\exp(-W_{rep}\sum_{m=1}^{P}(\theta_m - \theta_m^i)^2)\right] \quad (15)$$

where $J(\theta,P(j,k,l,m))$ is the computed dynamic objective function value (not the real objective function in the problem), $S$ is the total number of bacteria, $P$ is the number of variables to be optimized (embedded in each bacterium), $H_{rep}$, $W_{rep}$, $H_{att}$, and $W_{att}$ are user-defined parameters.

During reproduction, the healthy bacteria or the bacteria which are successful in securing a high degree of nutrients are let to reproduce asexually by splitting into two. Bacteria which do not manage to perform according to the specified criteria are eliminated from the group and thus not allowed to reproduce causing their genetic propagation (in this case their foraging strategies) to come to a halt. Due to this cycle, the amount of individual bacterium in the swarm remains constant throughout the execution of the BFA. Catastrophic events in an environment (such as a sudden change in physical/chemical properties or rapid decrease in nutrient content) can effect in death to a population of bacteria. Such events can cause bacteria to be killed and some to be randomly dispersed to different locations in the objective space. These events which are set to occur in the elimination/dispersal phase help to maintain swarm diversity to make sure the search operation is efficient. Figure 3 shows the workflow of the BFA technique. The pseudo-code for the BFA approach is provided below:

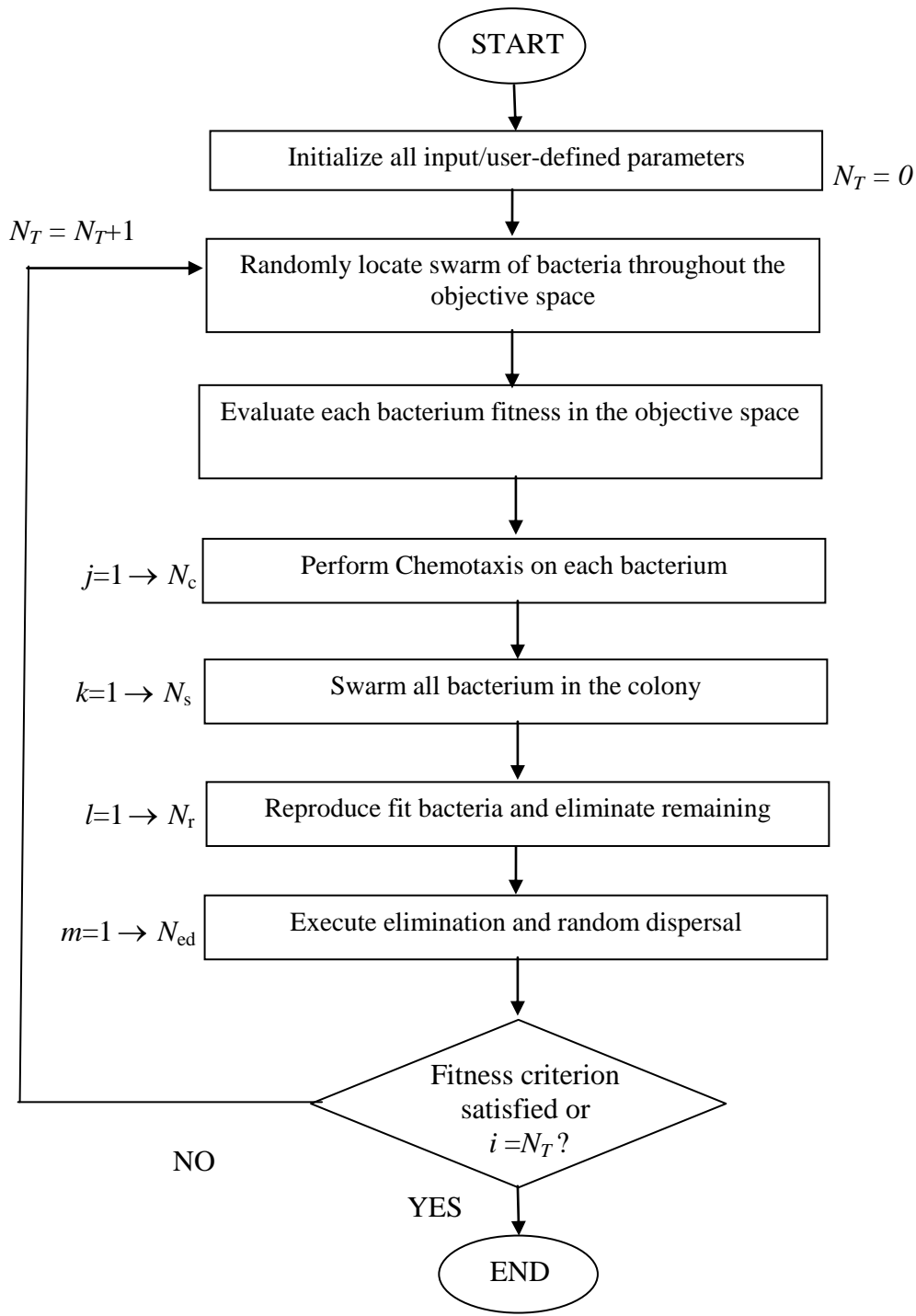

Figure 3: The workflow of the BFA

**START PROGRAM**
Initialize all input parameters ($S$, $P$, $H_{rep}$, $W_{rep}$, $H_{att}$, $W_{att}$, $N_T$, $N_c$, $N_r$, $N_s$, $N_{ed}$)
Generate a randomly located swarm of bacteria throughout the objective space
Evaluate bacteria fitness in the objective space
**For** $i=1 \rightarrow N_T$ **do**
    **For** $l=1 \rightarrow N_r$ **do**
        **For** $m=1 \rightarrow N_{ed}$ **do**
            **For** $j=1 \rightarrow N_c$ **do**
                **For** $k=1 \rightarrow N_s$ **do**
                Perform *chemotaxis* – bacterium swim and tumble until maximum fitness/loop cycle limit is reached
                Perform *swarming* – bacterium swarm until maximum fitness/loop cycle limit is reached
                **End For**
            **End For**
        **If** bacterium healthy/maximally fit **then** split and *reproduce*
        **Else** *eliminate* remaining bacterium
        **End For**
    Execute catastrophic *elimination* by assigning some probability of elimination to the swarm. Similarly *disperse* the remaining swarm randomly.
    **End For**
**End For**
**END PROGRAM**

## SIGMA DIVERSITY METRIC

The diversity measure used in this work is the sigma diversity metric (Mostaghim and Teich, 2003). The Sigma Diversity Metric (SDM) evaluates the locations of the solution vectors in the objective space relative to the sigma vectors. For lower dimensional objective spaces ($n < 3$), metrics that are based on spherical and polar coordinates could be used. However, as the dimensions increase beyond three ($n \geq 3$), the mentioned coordinate systems do not define the distribution of the solution vectors well (Mostaghim and Teich, 2005). In such scenarios, the SDM is highly effective for computing the solution distribution. To begin the computation of the SDM, two types of sigma lines would have to be constructed. First the sigma lines that represent the solution vectors, $\sigma'$ and the sigma lines that represent the reference lines, $\sigma$. The sigma lines that represent the solution vectors can be computed as the following:

$$\sigma'_k(ij) = \frac{f_i^2 - f_j^2}{\sum_{l=1}^{n} f_l^2} \qquad \text{such that} \qquad \forall i \neq j \qquad (16)$$

where $k$ denotes the index that represents the number of solution vectors, $i$, $j$ and $l$ denotes the index that represents the number of objectives and $n$ denotes the total number of objectives. Then the magnitude sigma $|\sigma'_k|$ is computed as follows:

$$\left|\sigma'_k\right| = \sqrt{\sum_{i=1}^{m}\sum_{j=1}^{m}\sigma'_k(ij)} \qquad (17)$$

Thus, for each line in the objective space (solution vector or reference line), there exists a unique sigma value. The central working principle is that the inverse mean distance of the solution vectors from the reference sigma vectors are computed. Since the reference sigma vectors are distributed evenly along the objective space, the inverse mean distance depicts the diversity of the solution spread. High values of the sigma diversity metric, indicates high uniformity and diversity in terms of the distributions of the solution vectors in the objective space.

## COMPUTATIONAL OUTCOME & ANALYSIS

In this work, all computational procedures (algorithms and metrics) were developed using the Visual C++ Programming Language on a PC with an Intel i5-3470 (3.2 GHz) Processor. The compromised solutions were obtained using the BFA and utilized for the construction of the Pareto frontier. The fuzzy MO design problem was converted to a scalarized aggregate single-objective form using the weighted sum framework. Hence for various scalarization, the compromised solutions to the MO problem were obtained. In this work, each Pareto frontier was constructed using a cumulative of 35 solution points. The fuzzified ambient temperature, $Z_a$ and insolation, $Z_b$ is shown in Figures 4 and 5 respectively:

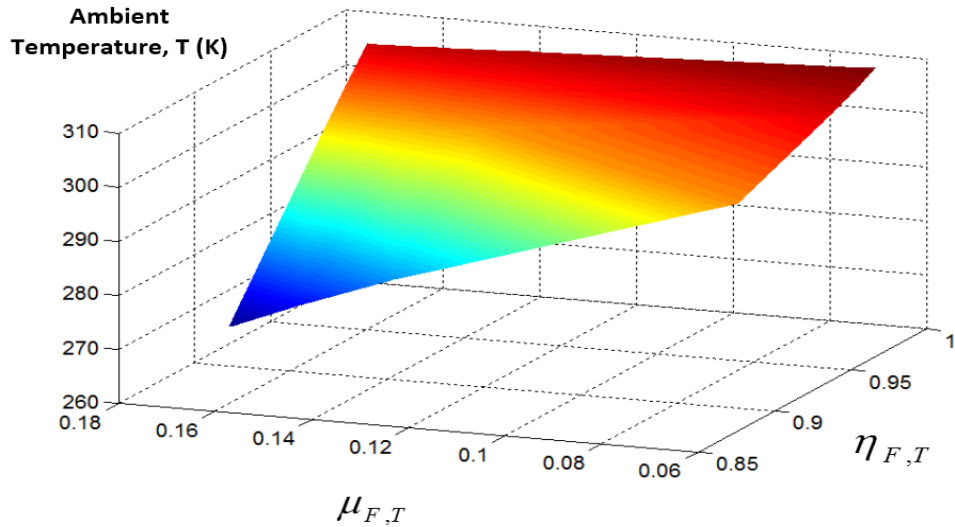

Figure 4: Fuzzy Ambient Temperature ($Z_a$) versus Membership grades

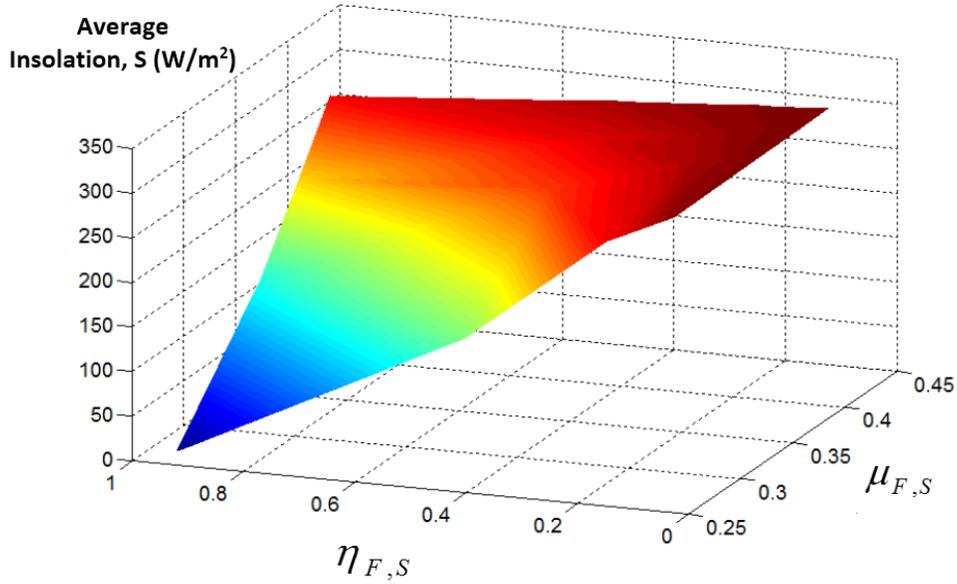

Figure 5: Fuzzy Insolation ($Z_b$) versus Membership grades

As in equations 3 and 4, the $\mu_{F,S}$ and $\mu_{F,T}$ are the primary membership grades while $\eta_{F,S}$ and $\eta_{F,T}$ are the secondary membership grades. The $S$ and $T$ subscripts denote the insolation and ambient temperature respectively. When performing type-2 fuzzy modeling, the number of membership grades are often large. Thus, data acquisition may become rather complicated as compared to data acquisition with fuzzy type-1 systems. To reduce the complexity of data acquisition, some approximate symmetry properties of the fuzzy membership grades are exploited. Thus, data acquisition could be performed by fixing the insolation membership grades and letting the ambient temperature membership grades vary. The other scenario being fixing the ambient temperature membership grade and varying the insolation grades. Due to the symmetrical properties of the membership grades, the final results should mirror each other in both scenarios. Evoking the symmetry property, the mapping of the objectives to the membership grades are represented in terms of membership grades of the ambient temperature while the membership grades of the insolation are let to vary in the following ranges:

$$\mu_{F,S} = [0.25264, 0.39913]$$
$$\eta_{F,S} = [0.02907, 0.92274] \tag{18}$$

The mapping of the power output, $f_1$ (kW) to various primary and secondary membership grades (of ambient temperature) are depicted in Figure 6:

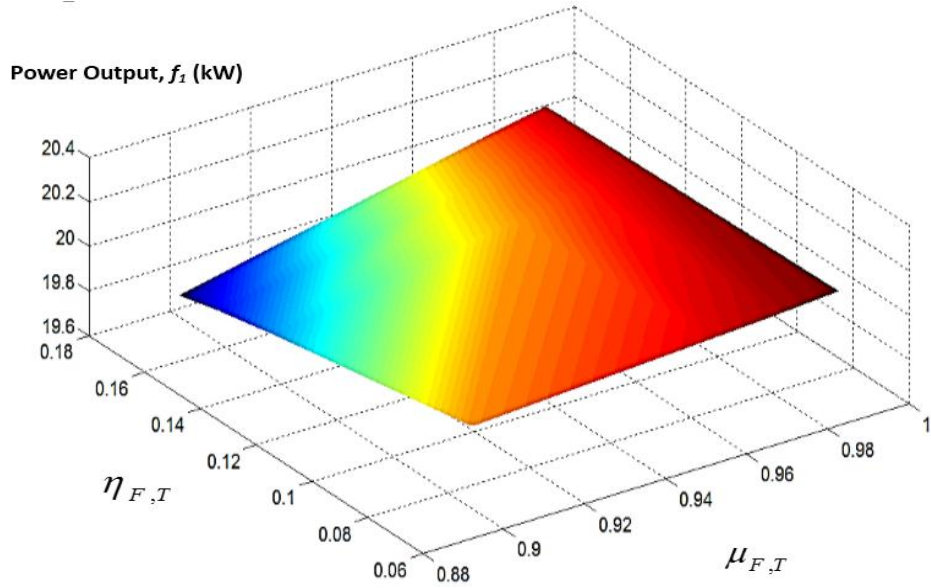

Figure 6: Power Output, $f_1$(kW) relative to Membership grades

Referring to Figure 6, the maximal variation of power outputs at all membership grades is 0.3671 kW. A maximum power output of 20.1267 kW was obtained at $\mu_{F,T} = 0.9871$ and $\eta_{F,T} = 0.0665$. The minimal power output at $\mu_{F,T} = 0.8967$ and $\eta_{F,T} = 0.1717$ was 19.7596 kW. Figure 7 and 8 show the mapping of the overall efficiency, $f_2$ (%) and the fiscal savings, $f_3$ (USD) respectively relative to the primary and secondary membership grades (ambient temperature):

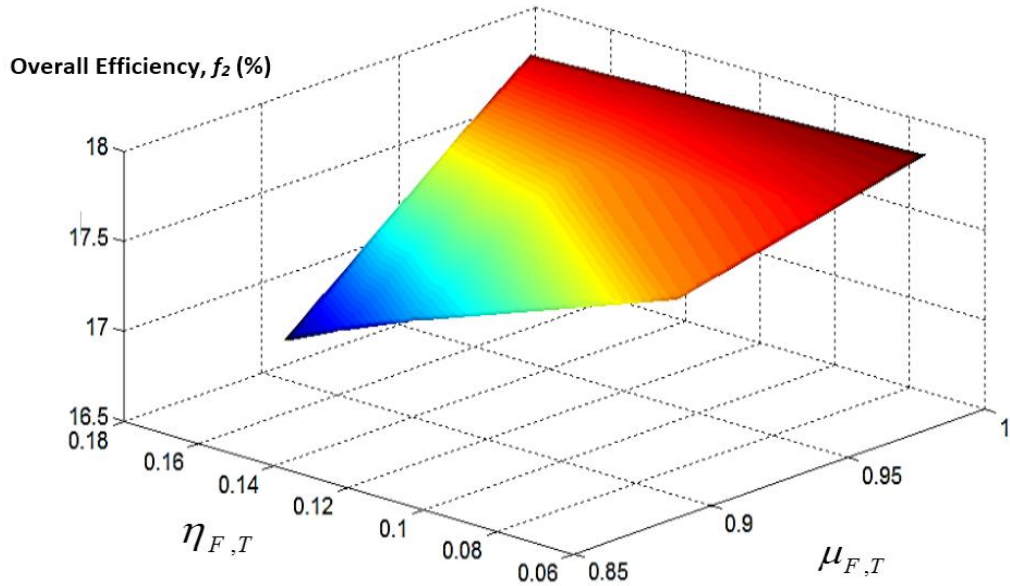

Figure 7: Overall Efficiency, $f_2$(%) relative to Membership grades

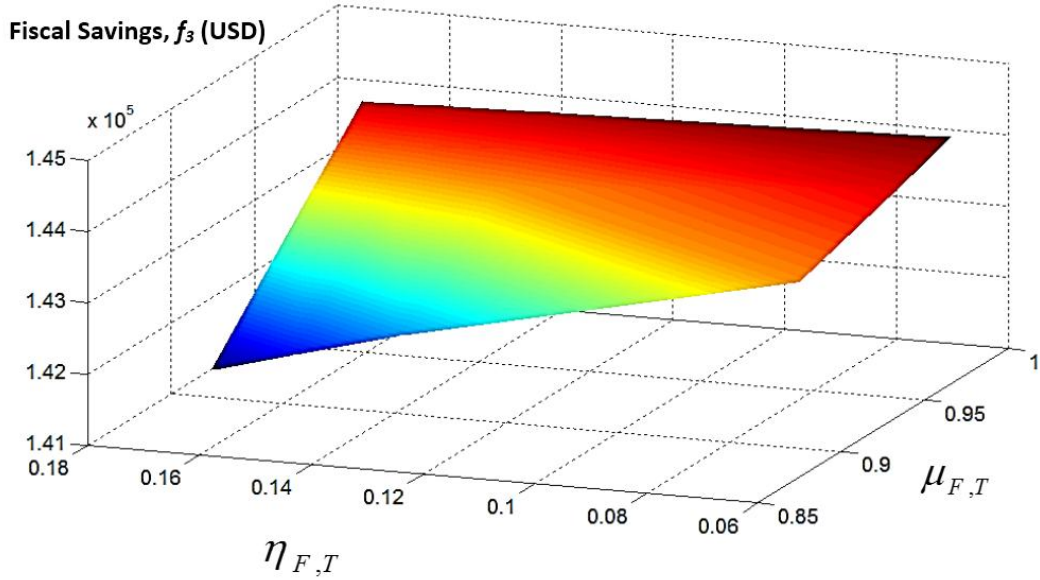

Figure 8: Fiscal Savings, $f_3$ (USD) relative too Membership grades

In Figure 7, the maximal overall efficiency of 17.9509% was obtained at $\mu_{F,T} = 0.9871$ and $\eta_{F,T} = 0.0665$ while the minimal overall efficiency of 16.7487% was obtained at $\mu_{F,T} = 0.8967$ and $\eta_{F,T} = 0.1717$. The highest variation in overall efficiency is 1.022%. Referring to Figure 8, the fiscal savings reaches the maximum of 144113 USD (at $\mu_{F,T} = 0.9871$ and $\eta_{F,T} = 0.0665$) with the minimal of 141434 USD ($\mu_{F,T} = 0.8967$ and $\eta_{F,T} = 0.1717$). The variation in fiscal savings with respect to the membership grades is 2679 USD. In Figures 6 – 8, it can be observed that except for the $f_3$ (fiscal savings), the variation in values of the objectives with respect to the membership grades are very small, $(f_1, f_2, f_3) = (0.3671\text{kW}, 1.022\%, 2679 \text{ USD})$. The design model presented in this chapter is for a single unit. In real-world applications, multiple units are conventionally utilized for stable power supply to large irrigation systems. Therefore, although the variations in the objectives are low, when projected to a larger scale such variations may compound producing a greater impact in terms of power output, system efficiency and fiscal savings. It should be noted, that even when considering a single-unit system, the variations in the membership grades (which spring from the noisy and uncertainty in insolation and ambient temperature) significantly affects the cost of the system (2679 USD).

For specific primary and secondary membership grades, three Pareto frontiers were selected based on the most optimal values of the objectives. The details on the membership grades are specified to construct these frontiers are given in Table 2:

*Table 2. Pareto frontiers and their membership grades*

| Description | $\mu_{F,T}$ | $\eta_{F,T}$ | $\mu_{F,S}$ | $\eta_{F,S}$ |
|---|---|---|---|---|
| *Frontier 1* | 0.8967 | 0.17169 | | |
| *Frontier 2* | 0.9565 | 0.15725 | 0.25264 - 0.39913 | 0.02907 - 0.92274 |
| *Frontier 3* | 0.9871 | 0.06648 | | |

Referring to Table 2, it is seen that the ambient temperature memberships are specified while the insolation memberships are left to vary in their ranges. This is done via the utilization of the symmetry properties in the membership grades. The individual solutions for various weights generated by the BFA were gauged and ranked based on the values of the aggregate objective function. Table 3 provides the ranked individual solutions for Frontier 1:

*Table 3. Individual Solution Rankings for Frontier 1*

| Description | | Best | Median | Worst |
|---|---|---|---|---|
| Objective Function | $f_1$ | 20.6787 | 20.6666 | 20.6466 |
|  | $f_2$ | 17.1771 | 17.1173 | 17.0407 |
|  | $f_3$ | 148004 | 147917 | 147774 |
| Decision Parameters | $x_a$ | 0.355846 | 0.354952 | 0.353833 |
|  | $x_b$ | 460.647 | 460.504 | 460.267 |
|  | $x_c$ | 677.284 | 674.784 | 671.6 |
|  | $x_d$ | 0.030827 | 0.030491 | 0.030037 |
| Noise Factors | $Z_a$ | 251.838 | 251.837 | 251.835 |
|  | $Z_b$ | 328.974 | 328.972 | 328.97 |
| Aggregate Objective | $F$ | 118407 | 44388.1 | 14794.9 |

In Table 1, the best individual solution was obtained at the weights (0.1, 0.1, 0.8) while the worst solution was attained at (0.6, 0.3, 0.1). The weight assignment for the median solution was (0.3, 0.4, 0.3). The scattered solutions in the objective space approximating the Pareto Frontier 1 is shown in Figure 9:

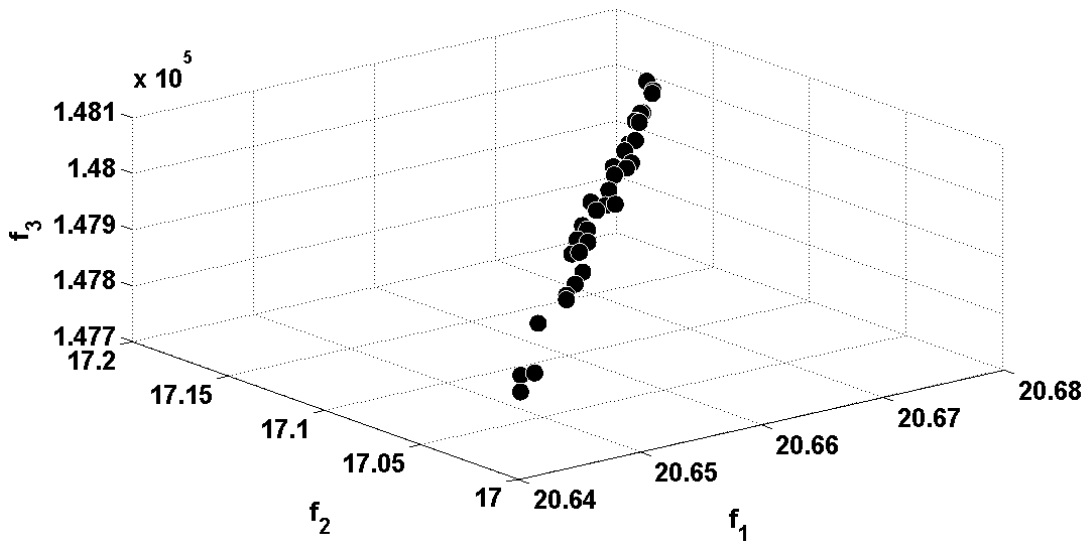

Figure 9: Approximation of Frontier 1

The ranked individual solutions along with their respective aggregate objective functions and parameters for Frontier 2 is given in Table 4:

*Table 4. Individual Solution Rankings for Frontier 2*

| Description | | Best | Median | Worst |
|---|---|---|---|---|
| Objective Function | $f_1$ | 20.9103 | 20.9103 | 20.8872 |
| | $f_2$ | 17.7455 | 17.7506 | 17.6432 |
| | $f_3$ | 149696 | 149696 | 149531 |
| Decision Parameters | $x_a$ | 0.354758 | 0.354881 | 0.353615 |
| | $x_b$ | 460.473 | 460.473 | 460.202 |
| | $x_c$ | 674.513 | 674.732 | 670.244 |
| | $x_d$ | 0.030421 | 0.030458 | 0.029886 |
| Noise Factors | $Z_a$ | 277.854 | 277.854 | 277.852 |
| | $Z_b$ | 328.972 | 328.972 | 328.969 |
| Aggregate Objective | $F$ | 119761 | 44921.7 | 14969.3 |

The best, worst and median individual solutions presented in Table 4 are associated with the weights ((0.1, 0.1, 0.8), (0.1, 0.8, 0.1) and (0.1, 9.6, 0.3) respectively. The graphical representation of the Pareto frontier 2 is depicted in Figure 10:

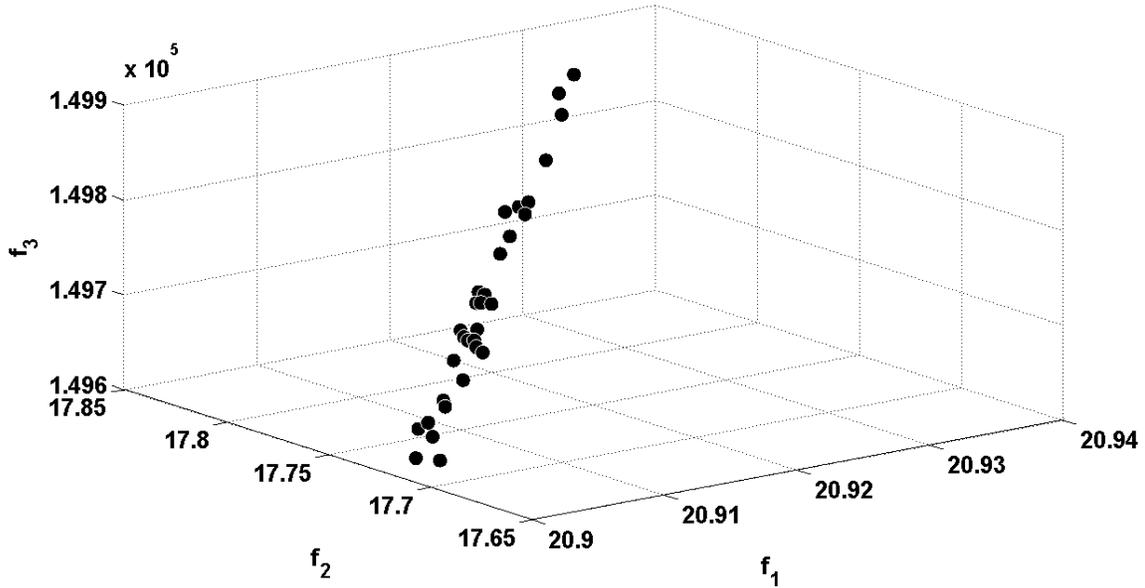

Figure 10: Approximation of Frontier 2

Table 5 gives the ranked individual solutions for frontier 3 along with its respective parameters, noise factors and objective values:

*Table 5. Individual Solution Rankings for Frontier 3*

| Description | | Best | Median | Worst |
|---|---|---|---|---|
| Objective Function | $f_1$ | 21.034 | 21.0524 | 21.0335 |
| | $f_2$ | 18.0657 | 18.1411 | 18.0577 |
| | $f_3$ | 150600 | 150731 | 150596 |
| Decision Parameters | $x_a$ | 0.354785 | 0.355775 | 0.354627 |
| | $x_b$ | 460.435 | 460.65 | 460.428 |
| | $x_c$ | 674.234 | 677.372 | 673.892 |
| | $x_d$ | 0.030391 | 0.030815 | 0.030349 |
| Noise Factors | $Z_a$ | 291.26 | 291.262 | 291.26 |
| | $Z_b$ | 328.972 | 328.974 | 328.971 |
| Aggregate Function | $F$ | 120484 | 45232.6 | 15077.3 |

The best and worst individual solutions are associated with the weights (0.1, 0.1, 0.8) and (0.5, 0.4, 0.1) respectively. The median solution has the weights (0.2, 0.5, 0.3). The Pareto Frontier 3 is shown in Figure 11:

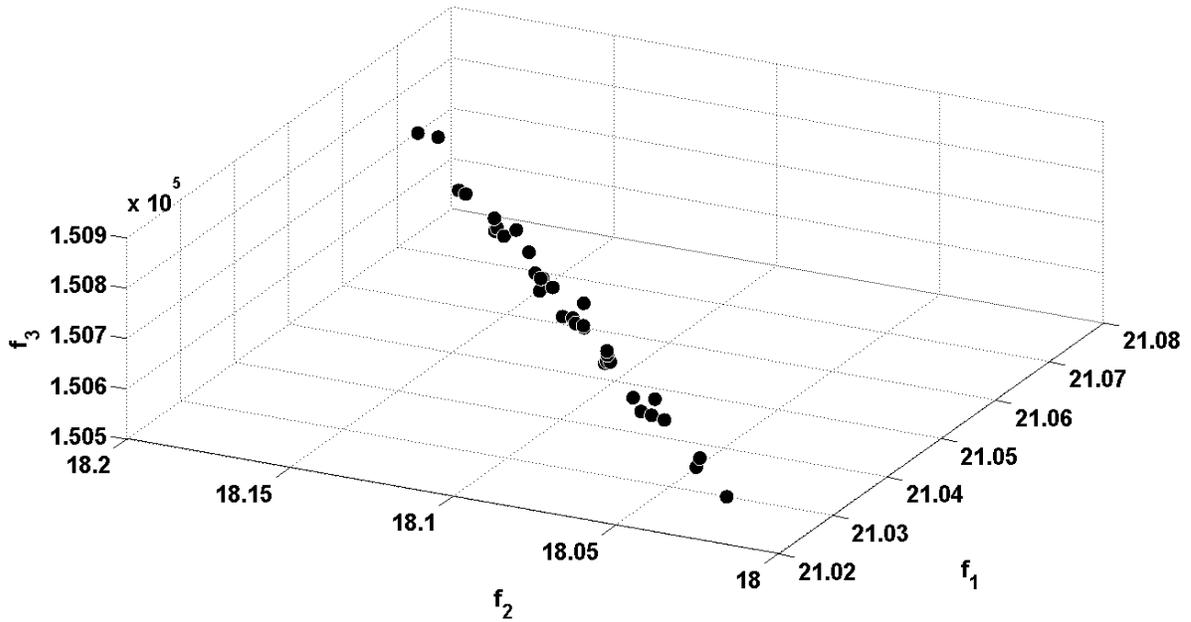

Figure 11: Approximation of Frontier 3

The three frontiers generated by the BFA technique was obtained at various weights for three different membership grades (refer to Table 2). In this chapter, the overall dominance of the Pareto frontier is measured by taking the average value of the aggregate objective function, $F$ across the entire frontier. Frontier 3 outranks Frontier 2 followed by Frontier 1. The average value of the objective function across the entire frontier for Frontier 1, 2 and 3 are: 50305.86, 50911.64 and 51237.71 respectively. Similarly in Tables 3-5, it can be observed that the value of the aggregate objective function for the best individual solution follows a similar trend as compared to the entire Pareto frontiers in terms of ranking. The Pareto frontier were also gauged using the SDM as presented in the previous section. The diversity levels obtained for each of the frontiers are presented in Figure 12:

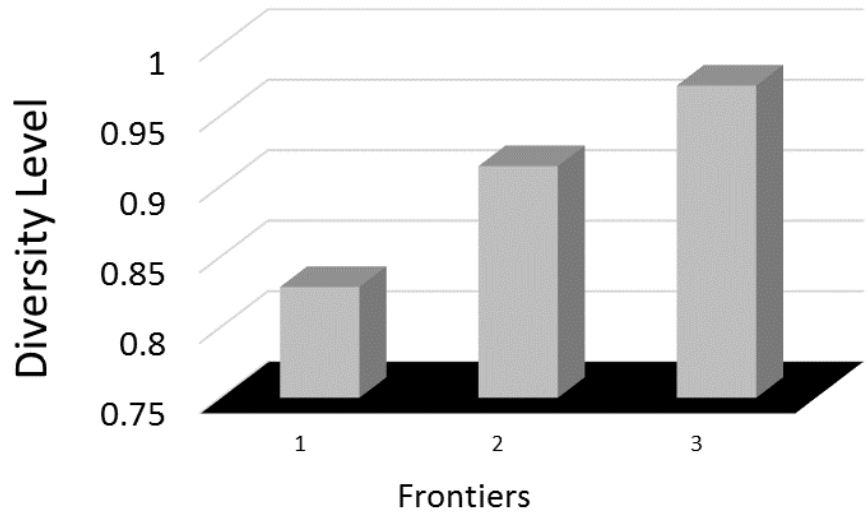

Figure 12: Diversity Levels for the Pareto Frontiers

In MO optimization, the capability of the solution technique to continuously produce solutions with high diversity is crucial. This is so that the solution technique (algorithm) does not stagnate at certain locations in objective space or the local optima. Such stagnations may inhibit the algorithm from exploring other regions in the objective space which may contain other local optima which are much closer to the global optima. Therefore, measuring the diversity of the solution spread across the Pareto frontier provides the user/decision maker with information about the performance of the algorithm during its search operations. The BFA employed in this work seem to generate a highly diverse solution spread which constructs the Pareto Frontier 3 (Figure 12). Frontier 2 has lower diversity that Frontier 3 followed by Frontier 1 which has the lowest diversity as compared to all the frontiers generated using the BFA technique.

Since diverse solution generation provides the algorithm with the ability to overcome stagnation, the degree of frontier dominance closely relates to the diversity of the solution spread across the concerned frontier. In this work, it is seen that the diversity levels follow the ranking of the dominance measured by using the average aggregate objective function value. Therefore, the BFA approach generated diverse solutions during its execution which resulted in the construction of highly dominant Pareto frontiers.

One of the most vital issues in green energy engineering is the design of low-cost and efficient systems which could match (or surpass) its fossil-fuel counterparts. However, these design efforts are often faced with challenges due to the system's interactions with noisy and uncertain environments. Identification and classification of the type of uncertainty is essential for the conception of suitable design frameworks. Using such frameworks, uncertainty-tolerant systems could be optimally designed and manufactured. In view of this idea, the data consisting of environmental factors presented in Table 1 could be seen to have two levels of division: monthly average data and annual average. Thus, modeling this data using a fuzzy type-1 system is clearly inadequate and may lead to inaccurate approximations to the solar-powered irrigation system's design characteristics. It could be observed that the reformulation of the problem in a fuzzy type-2 MO programming setting offers various new information such as the system tolerance limits and its behaviour when under the influence of uncertainties. The three frontiers in Figures 9-11 were extracted because they contain the most significant variations in the target objectives when exposed to the uncertainties in the environmental factors. The maximal variation in the power output when faced with uncertainties (represented by the membership grades of the frontiers) is 0.3548 kW. The highest variation of the overall efficiency and the fiscal savings considering the mentioned uncertainties are 0.8806% and 2592 USD respectively. Therefore, in a large-scale solar irrigation system, uncertainties in the environmental factor (specifically: insolation levels and ambient temperature) does significantly affect the design properties of these systems.

The BFA technique like most metaheuristic approaches is a stochastic optimization technique. Therefore due to tis random nature, multiple runs are required during execution for result consistency. In this work, each individual solution is obtained after 5 program executions which adds up to 175 runs per frontier. The parameter setting of the BFA were heuristically determined and were not varied in this work. The parameter setting for the BFA is as in Table 6:

*Table 6: Initial Parameters for the BFA*

| Parameters | Values |
| --- | --- |
| Total Iteration, $N_T$ | 200 |
| Population Size, $P$ | 25 |
| Swimming Loop limit, $N_S$ | 5 |
| Repellent Signal Width, $W_{rep}$ | 10 |
| Attractant Signal Width, $W_{att}$ | 0.2 |
| Repellent Signal Height, $H_{rep}$ | 0.1 |
| Attractant Signal Height, $H_{att}$ | 0.1 |
| Reproduction limit, $N_r$ | 5 |
| Elimination limit, $N_e$ | 5 |

Referring to Figure 3, it can be observed that the BFA has one overall loop plus four primary loops: chemotaxis, swimming, reproduction and elimination/dispersal. Due to its rigor in solution discrimination portrayed by the cascaded loops, the BFA has high algorithmic complexity. The BFA code employed in this work contains multiple subroutines to account for the BFAs complexity. This significantly influences the computational time of the BFA during execution. The BFA program employed in this work takes an average of 4.217 seconds to compute each individual solution per run with an average of 1500 iterations. Therefore to construct the entire frontier for this application, the BFA takes approximately 12.3 minutes. Besides the algorithmic complexity of the BFA, the objective space of the solar-powered irrigation problem also contributes to the computational effort and time. This is because, the objective space in this problem contains multiple local optima that misleads the algorithm into stagnation. To overcome these local optima and proceed with the computation, the BFA takes additional computational time.

Another important feature of a MO optimization approach is the fitness assignment. The fitness assignment in the computational approach affects the way the algorithm is driven during the search process. In this work, all three target objectives are maximized and no objectives are minimized. Therefore, the aggregate objective function is used as a fitness criterion for the BFA algorithm during the search. The fitness of swarm is considered improved if the aggregate objective function is maximized during the current iteration as compared to the previous iteration. It is important to note that although the diversity metric is employed to measure the diversity of the solutions in the frontier, the diversity metric was not assigned as a fitness criterion in the algorithm. The diversity metric was employed in an offline manner from time to time to ensure that the algorithm is producing a diverse population during execution.

Diversity preservation is crucial when it comes to dealing with MO programming algorithms. In this work, two diversity preservation mechanism were utilized. The first one as mentioned previously is the offline implementation of the diversity metric. The second approach is the in-built diversification mechanism of the BFA. The in-built mechanism of the BFA consists of the two subcomponents which are the initial randomization and the random dispersal components (see Figure 3). The initial randomization component, randomly locates each of the bacterium at various locations in the objective space prior to the search. The random dispersal component, spreads the bacteria which survived elimination randomly throughout the objective space. This random spreading diversifies the bacteria population preparing it for the next program iteration. These two subcomponents randomizes the BFA during execution such that the bacterium population is sufficiently diverse during the search.

In the BFA, solution elitism could be readily ensured by the process of elimination (refer to Figure 3). In this process, bacteria which does not perform improvements based on the fitness assignment (which in this case is the maximization of the aggregate objective function) is eliminated from the population pool. This elitism mechanism ensures that only the elite bacteria is allowed to swarm during the consequent program cycles. Throughout the execution, the BFA algorithm performed in a stable manner and managed to successfully converge to a solution during each run. Besides, none of the fuzzy constraints were violated by solutions generated by the BFA. Therefore, the BFA reliably produces feasible solutions for all the associated weights. This in effect enables the BFA to capture solutions in the objective space which are close to the global optima.

## CONCLUSIONS

The MO design of the solar-powered irrigation system was completed using type-2 fuzzy modeling to represent uncertain environmental factors. Using the BFA in tandem with the weighted-sum approach, dominant Pareto frontiers were constructed using individual solutions (with various weights). The dominance of the three generated Pareto frontiers were evaluated and ranked. The design problem was successfully reformulated using type-2 fuzzy logic to depict uncertainty in the environmental data (as shown in Table 1). As seen in Tables 3-5, new optimal individual solutions were obtained. In addition, information regarding the effects of uncertainties on the variations in the target objectives were ascertained and analyzed. This knowledge would prove to be very useful for designers dealing with solar-powered irrigation systems which operate in uncertain (or noisy) environments. In addition, using the diversity metric and the aggregate objective function, the quality of the solutions as well as the behaviour of the algorithm was analyzed in detail. The operations and the mechanisms of the BFA related to the solar-powered irrigation problem is presented. Characteristics such as algorithmic complexity and execution time was also presented and discussed.

It can be concluded that although the BFA consumes high computational resources, it compensates in terms rigor in solution scrutiny. This in turn provides the engineer/decision maker with high quality solutions. Based on the results obtained in this work, it can be seen that if a multi-unit (large-scale) solar-powered irrigation system were to be designed for operation in an uncertain environment, its key properties such as efficiency, cost and power output would be greatly affected.

## FUTURE RESEARCH DIRECTIONS

In the future, to address the issue of algorithmic uncertainty, the BFA should be implemented to this design problem however by varying its initialization parameters. This way, the optimal parameters could be identified for solving this design problem with uncertain environmental factors. Besides, other conventional fuzzy-based approaches (Vasant *et al*., 2010; Ganesan *et al*., 2014) could be upgraded to a type-2 fuzzy framework and implemented to this problem. Other forms of metaheuristic approaches such using evolutionary strategies could be implemented to this design problem (Ganesan *et al*., 2015). Other measurement metrics such as the hypervolume indicator could be employed for gauging the degree of dominance of the generated Pareto frontiers (Zitzler and Thiele, 1998). A study on rigorously determining the fuzzy parameters (equation (3)) could be done to obtain better modeling accuracy. This would be very useful for future researchers attempting to optimize the design of this system.

## ACKNOWLEDGEMENTS

The authors are very grateful to the reviewers for their efforts in providing useful and constructive insights that has been overlooked by the author during the preparation of the manuscript. The authors would also like to thank Universiti Teknologi Petronas for their support during progress of this work.

Special thanks to our friends, colleagues and students who continuously inspire and support us throughout the phases of our research.

Ganesan, T., Vasant, P. and Elamvazuthi, I., (2014), Hopfield neural networks approach for design optimization of hybrid power systems with multiple renewable energy sources in a fuzzy environment, Journal of Intelligent and Fuzzy Systems, Vol 26 (5), pp 2143-2154.

Ganesan, T., Elamvazuthi, I. and Vasant, P., (2015), Multiobjective Design Optimization of a Nano-CMOS Voltage-Controlled Oscillator Using Game Theoretic-Differential Evolution, Applied Soft Computing, doi:10.1016/j.asoc.2015.03.016.

Zitzler, E. & Thiele, L., (1998), Multiobjective Optimization Using Evolutionary Algorithms - A Comparative Case Study, In Conference on Parallel Problem Solving from Nature (PPSN V), 292–301.

## ADDITIONAL READING

Beume, N., Naujoks, B., & M. Emmerich. (2007), SMS-EMOA: Multiobjective selection based on dominated hypervolume. *European Journal of Operational Research*, 181(3), 1653–1669.

Igel, C. Hansen, N. & Roth, S., (2007), Covariance matrix adaptation for multi-objective optimization. Evolutionary Computation, 15(1), 1–28.

Knowles, J. & Corne, D. (2003), Properties of an adaptive archiving algorithm for storing nondominated vectors. IEEE Transactions on Evolutionary Computation, 7(2), 100–116.

Zelinka I., (2002), Analytic programming by Means of Soma Algorithm. Mendel '02, In: *Proceeding of the 8th International Conference on Soft Computing Mendel'02, Brno*, Czech Republic, 93-101.

Varacha, P., (2011), 'Neural Network Synthesis via Asynchronous Analytical Programming', *Recent Researches in Neural Networks, Fuzzy Systems, Evolutionary Computing and Automation*

Zelinka I. & Oplatkova Z., (2003), Analytic programming – Comparative Study. CIRAS'03, *Second International Conference on Computational Intelligence, Robotics, and Autonomous Systems,* Singapore.

Oplatkova Z., & Zelinka I. (2006), Investigations using Artificial Ants using Analytical Programming, *Second International Conference on Computational Intelligence, Robotics, and Autonomous Systems,* Singapore.

Oplatkova Z., & Zelinka I. (2007), Creating evolutionary algorithms by means of analytic programming - design of new cost function. In *ECMS 2007, European Council for Modelling and Simulation*, 271-276.

Koza, J.R., (1992), Genetic Programming: On the Programming of Computers by means of Natural Selection, *MIT Press*, USA.

Ganesan, T., Elamvazuthi, I., K.Z.K., Shaari and Vasant, P., (2013), Hypervolume-driven analytical programming for solar-powered irrigation system optimization, Nostradamus 2013: Prediction, Modeling and Analysis of Complex Systems, pp. 147-154.

## KEY TERMS & DEFINITIONS

- Multiobjective optimization

Optimization problems which are represented with more than one objective functions.

- Metaheuristics

A framework consisting of a class of algorithms employed to find good solutions to optimization problems by iterative improvement of solution quality.

- Solar-powered irrigation system

Systems which harness solar energy to power irrigation pumps.

- Swarm Intelligence

A group or collective behavior of natural or artificial individuals in a system towards some target.

- Bacteria Foraging Algorithms (BFA)

A type of swarm algorithm that uses the dynamics of bacteria foraging to search for optimal solutions in the objective space.

- Type-2 Fuzzy Logic

Type-2 fuzzy logic is generalization of the conventional or type-1 fuzzy logic for handling high levels of uncertainty.